%% file: main.tex
\documentclass[letterpaper]{article} 
\usepackage{aaai2026}  
\usepackage{times}  
\usepackage{helvet}  
\usepackage{courier}  
\usepackage[hyphens]{url}  
\usepackage{graphicx} 
\usepackage{natbib}  
\usepackage{caption} 
\usepackage{cleveref}

\usepackage{algorithm}
\usepackage{algorithmic}
\usepackage{booktabs}
\usepackage{makecell}

\usepackage{subfig}

\usepackage{xcolor}

\newcommand{\answerYes}[1]{\textcolor{blue}{#1}} 
\newcommand{\answerNo}[1]{\textcolor{teal}{#1}} 
\newcommand{\answerNA}[1]{\textcolor{gray}{#1}}

\usepackage{newfloat}
\usepackage{listings}
\DeclareCaptionStyle{ruled}{labelfont=normalfont,labelsep=colon,strut=off} 
\floatstyle{ruled}
\newfloat{listing}{tb}{lst}{}
\floatname{listing}{Listing}
\usepackage{xurl}

\title{Misinformation Span Detection in Videos via Audio Transcripts}
\author {
    Breno Matos\textsuperscript{\rm 1,}\textsuperscript{\rm 2}\thanks{Work done while the author was at the Federal University of Minas Gerais.},
    Rennan C. Lima\textsuperscript{\rm 2,}\textsuperscript{\rm 4},
    Savvas Zannettou\textsuperscript{\rm 3},\\
    Fabrício Benevenuto\textsuperscript{\rm 2},
    Rodrygo L.T. Santos\textsuperscript{\rm 2},
}
\affiliations {
    \textsuperscript{\rm 1}Max Planck Institute for Informatics\\
    \textsuperscript{\rm 2}Federal University of Minas Gerais\\
    \textsuperscript{\rm 3}TU Delft\\
    \textsuperscript{\rm 4}Kunumi Institute\\
    bdesousa@mpi-inf.mpg.de, s.zannettou@tudelft.nl, \{rennancordeiro, fabricio, rodrygo\}@dcc.ufmg.br
}

\begin{document}

\maketitle

\begin{abstract}
Online misinformation is one of the most challenging issues lately, yielding severe consequences, including political polarization, attacks on democracy, and public health risks. Misinformation manifests in any platform with a large user base, including online social networks and messaging apps. It permeates all media and content forms, including images, text, audio, and video. Distinctly, video-based misinformation represents a multifaceted challenge for fact-checkers, given the ease with which individuals can record and upload videos on various video-sharing platforms. Previous research efforts investigated detecting video-based misinformation, focusing on whether a video shares misinformation or not on a video level. While this approach is useful, it only provides a limited and non-easily interpretable view of the problem given that it does not provide an additional context of \emph{when} misinformation occurs within videos and \emph{what} content (i.e., claims) are responsible for the video's misinformation nature. 

In this work, we attempt to bridge this research gap by creating two novel datasets that allow us to explore misinformation detection on videos via audio transcripts, focusing on identifying the span of videos that are responsible for the video's misinformation claim (\emph{misinformation span detection}). We present two new datasets for this task, both containing false claims and the video moment in which they appear. We transcribe each video's audio to text, identifying the video segment in which the misinformation claims appears, resulting in two datasets of more than 500 videos with over 2,400 segments containing annotated fact-checked claims. Then, we employ classifiers built with state-of-the-art language models, and our results show that we can identify in which part of a video there is misinformation with an F1 score of 0.68. To assist the research community in future research endeavors focusing on misinformation span detection, we make publicly available our annotated datasets that includes false claims and the video spans that these false claims appear in videos. We also release all transcripts, audio and videos. 
\end{abstract}

\input{sections/introduction}

\input{sections/related_work}

\input{sections/methodology}
\input{sections/results}

\input{sections/limitations}
\input{sections/applications}
\input{sections/conclusions}
\input{sections/future-work}

\input{sections/ack}

\bibliography{aaai2026}

\input{sections/paper-checklist}

\input{sections/appendix}

\end{document}

%% file: sections/introduction.tex
\section{Introduction}


Misinformation is one of the most challenging problems in our society in recent years. The problem manifests itself in many different ways, for instance, usually working as an engine for campaigns that promote attacks on democracy, political polarization and radicalization~\cite{Shao2018,bessi2016social}, affecting the debate across multiple issues, including climate change~\cite{lutzke2019priming} and health, as evidenced during the COVID-19 pandemic, with the spread of anti-vaccination misinformation~\cite{qanon-savvas}.
Consequently, initiatives to tackle misinformation gained relevance, evidenced, for example, by the nomination of the International Fact-Checking Network (IFCN) to the Nobel Peace Prize in 2021~\cite{ifcnnobel}.

\begin{figure}
    \centering
    \includegraphics[width=0.8\columnwidth]{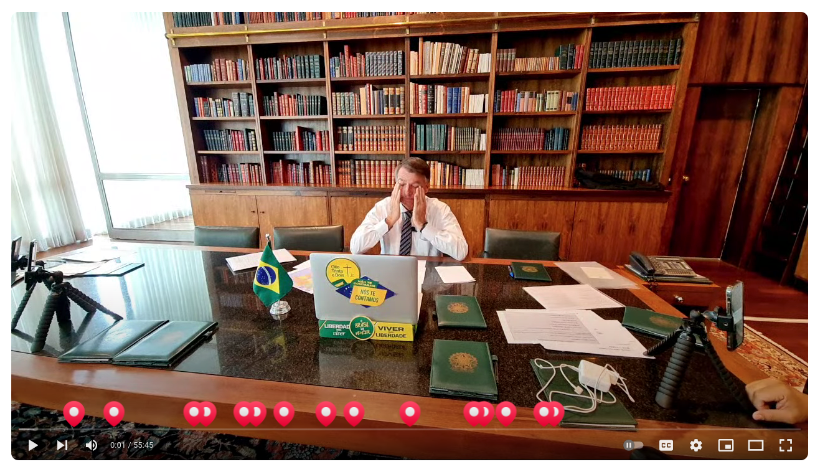}
    \caption{Example of a fact-checked video with pointers to misinformative segments}
    \label{fig:bolsonaro-fact-checked}
\end{figure}


One of the main challenges in combating misinformation lies in the complexity of digital platform environments and the various forms in which it can arise. For example, misinformation can be launched through websites that appear reliable sources of information but are, in reality, dedicated to disseminating misinformation, often with political motivations~\cite{budak}.
Moreover, misinformation can manifest through different formats, including memes, images, and content shared across social networks, specialized groups, and messaging platforms like WhatsApp \cite{zapmonitor1, zapmonitor2} and Telegram \cite{telegrammonitor1}. It spreads through various mediums, encompassing audio~\cite{audiomisinformation1}, video~\cite{facemanipulation, deepfake}, images~\cite{image3,image4}
, and text-based content~\cite{TWITTER4,textmisinformation1}.

Misinformation in video content represents a particularly complex problem due to the massive amount of videos uploaded daily on platforms like YouTube and TikTok. In a single day, YouTube receives a volume of user-generated videos equivalent to 720,000 hours\footnote{\url{https://www.globalmediainsight.com/blog/youtube-users-statistics/\#stat}}. 

Fact-checking agencies cannot keep up with the rapid spread of online misinformation without tools that facilitate journalists to identify content that is worth fact-checking. Additionally, content moderation in videos is a growing concern for platforms such as YouTube and TikTok, especially with novel regulations, such as the Digital Services Act (DSA), which forces platforms to remove content that is against their terms and also provide transparency about the moderation process. 

Thus, this scenario calls for automated detection methods of misinformation in videos. However, differently from the task of detecting if a textual claim or image is fake, a video can contain hours of speech and become a very labor-intensive task. 
Despite their undeniable importance, previous research focused on detecting whether a video shares misinformation or not on a video level~\cite{younicom, misinfordetectvideos, medicalvideos}. While this approach is useful, it only provides a limited and non-easily interpretable view of the problem given that it does not provide an additional context of \emph{when} misinformation occurs within videos and \emph{what} content (i.e., claims) are responsible for the video's misinformation nature.



In this work, we propose two novel datasets to allow the research community to approach the problem of misinformation span detection in videos, which involves determining the specific segments of a video where misinformation is present. For example, Figure~\ref{fig:bolsonaro-fact-checked} depicts a real 55-minute-long video, which was fact-checked by specialists who pointed out 16 misinformative claims (for an illustrative purpose, we marked with red dots the segments in which the false claims are made). Our effort in this work evaluates the feasibility of automatically spotting the segments of the videos where these false claims appear. To do it, we used a methodology based on a three-step approach. 

First, we gathered two datasets of videos verified by the fact-checking agency, namely Aos Fatos\footnote{https://www.aosfatos.org/},  which is part of the International Fact-Checking Network (IFCN) and one of the most prestigious fact-checking agencies in Brazil. Both datasets contain videos and a set of false claims made in the video. The first dataset contains 538 videos featuring Brazil's former president, Jair Bolsonaro, throughout his 4-year term. The second dataset comprises 78 videos containing electoral fraud claims made by voters during the 2022 Brazilian presidential election. Our second step consists of extracting textual transcripts from these videos and annotating the time in which each false claim appears in order to identify which segments of each video contain misinformation. Finally, in our last step, we set the first benchmark for this task through different evaluation scenarios, testing multiple classification approaches using modern language models in order to investigate the feasibility of differentiating misinformative and non-misinformative segments.

Our evaluation results indicate the feasibility of automated misinformation span detection in videos, pointing to valuable directions for developing tools that can assist fact-checkers and moderation in social media platforms. 
More importantly, by making this resource available, we open a whole new avenue of research possibilities, enabling the investigation of a relevant new problem, misinformation span detection, which can be explored with many different machine learning approaches, including potential multi-modal strategies.

Summarizing, our key contributions are:

\begin{itemize}
    \item To the best of our knowledge, we firstly approach  misinformation span detection in videos. We hope our methodology and results might offer not only guidance for future research in the theme, but also a baseline for comparison. Our results show that automatic detection is feasible for this task, though there remains scope for further improvement. Finally and more importantly, we hope our work can inspire future tools able to mitigate the misinformation problem in practice. 
    
    \item We publicly release two datasets containing 538 and 77 videos, annotated with timing in the videos in which 2,355 false claims and 78 false claims happen, respectively. To the best of our knowledge, this are the first datasets of their kind. We believe these datasets are  valuable resources for the research community.

\end{itemize}

%% file: sections/related_work.tex
\section{Related Work}\label{rw:misinforsocialmedia}

Social media platforms enabled much faster communication between users and increased the speed of information spread in general. However, this phenomenon also facilitated the spread of online misinformation, prompting platforms and researchers to present solutions to this problem.

Among all forms of misinformation, video is one of the most challenging, as it usually requires more processing power than, for instance, text, with previous works having proposed mitigation efforts for the issue. \citet{younicom} propose a dataset of conspiracy videos on YouTube and a pipeline to detect such videos. However, they perform classification at a video level, not pointing to where the conspiracy claims are made. \citet{medicalvideos} propose a similar approach for medical videos, also providing a dataset of annotated YouTube videos, but using an SVM-based classifier for their experiments.

Other works on misinformative videos focus on the platforms where they were uploaded, such as the work proposed by \citet{videomisinformation1}, which  audits YouTube and evidences how their recommendation systems can induce users to misinformative filter bubbles and grounding the need for more automated tools for misinformation detection on videos.

Additional works focus on manipulated videos: \citet{facemanipulation} focus on deceptive face manipulation on videos, also referred to as deepfakes, a form of misinformation built through synthetically generated media. Similarly, \citet{deepfake} centers on investigating whether deepfake detection methods proposed in the literature generalize to real-world deepfakes.

Other studies focus on short videos specifically: \citet{videomisinformation2} investigate misinformative videos about COVID-19 on TikTok by leveraging captions and video components to propose a classification approach. \citet{fakesv} also focuses on short video fake news and builds a dataset by crawling Chinese fact-checking portals, providing a baseline for binary multimodal detection of fake news videos' detection.

Another important work on misinformative videos was presented by \citet{misinfordetectvideos} where the authors propose a framework to classify videos into misinformation and non-misinformation, analyzing 2125 videos containing information about the vaccines controversy, the 9/11 conspiracy, chem-trails, the moon landing, and flat earth. However, like \cite{younicom,medicalvideos,videomisinformation2,videomisinformation1,fakesv}, they focus on video-level binary classification, without identifying where misinformation appears within the video.

In contrast to previous efforts, we introduce a novel dataset addressing the relevant yet under-explored problem of misinformation span detection in videos. Specifically, we set ourselves apart from previous works limited to the binary classification of videos containing misinformation. Our work also differs from previous ones that are limited to short videos. In summary, we propose a general approach to misinformation detection that can be used for videos of varying lengths while identifying which section of the video presents misinformative content, a task we frame as misinformation span detection.

%% file: sections/methodology.tex
\section{Problem Description}

Our work focuses on the \emph{misinformation span detection} task.
The objective of this task is the detection of spans that make a piece of content misinformative.\footnote{Our task is analogous to the Toxic Spans Detection task presented by Pavlopoulos et al.~\cite{pavlopoulos2021semeval}.} 
In other words, we aim to detect whether a piece of content is misinformative and, in particular, which spans of the content are responsible for the content's misinformative nature.
For instance, a long interview video may be considered misinformative due to brief segments in which false claims are made. 
Identifying these spans of false claims is paramount as it can assist fact-checkers and social media operators in providing the necessary context (e.g., warning labels) at the exact time of appearance of the false claims.

\section{Datasets}\label{sec:dataset}
Although finding mis/disinformation in videos is greatly important, previous work lacks sufficient data for the task of misinformation span detection. In this light, we build two novel datasets for the task: 1) BOL4Y and 2) EI22, further discussed below.

\subsection{BOL4Y dataset}\label{sec:firstdataset}
To build our first dataset, henceforth referred to as BOL4Y, we leverage a list of false claims made by Jair Bolsonaro, Brazil's former president. AosFatos, one of Brazil's biggest fact-checking agencies, compiled a list of 6,685 claims through \textbf{Bol}sonaro's \textbf{4-Year} presidential term~\cite{aosfatos_bolsonaro_declaracoes}. These claims come from multiple sources, such as interviews, written social media posts, and videos that Bolsonaro shared. 
Each fact check contains the following data:
\begin{itemize}
    \item \textbf{Claim:} A sentence that summarizes the false claim.
    \item \textbf{Fact-check:} Fact-check produced by AosFatos' journalists.
    \item \textbf{Broad theme:} The theme and broad topic of the claim, such as infrastructure or the COVID-19 pandemic.
    \item \textbf{Repetition count:} The number of times Bolsonaro made that claim on other occasions, including the dates for each occurrence.
    \item \textbf{Source:} The link to the source (e.g., social media post) that includes the false claim. Although most claims have repetitions throughout Bolsonaro's presidency, AosFatos only lists a source for one of those occurrences. Also, it includes the category of the source (e.g., interview, live stream, etc.). 
    \item \textbf{Media repercussion:} Links to other media websites that published a news piece about the claim.
\end{itemize}

We created our dataset by scraping AosFatos' website in March 2023, collecting data for 6,685 claims from 1,595 unique sources, which vary and include, for example, news pieces from major outlets, posts on social media, and official declarations on governmental websites. Then, we specifically focused on claims with video-based sources, primarily from social media platforms like YouTube, Facebook, TikTok, and occasionally from news websites.
Then, we visited the sources and downloaded the videos, obtaining a set of 525 videos.
Also, we note that for 121 claims, AosFatos did not provide a link to the source. 
However, they provide the transcript of the video that comes from AosFatos' transcription service, Escriba.~\footnote{\url{https://escriba.aosfatos.org/en/}}
We complement this dataset with these readily available textual transcripts.
Overall, this dataset includes 525 videos sharing false claims and 121 textual transcripts (corresponding to videos sharing false claims) that were obtained from AosFatos' transcription service.

\subsection{EI22 dataset}\label{sec:secondataset}

We also release a second expert-annotated dataset gathered by AosFatos. AosFatos fact-checked a set of videos posted on YouTube and privately shared them with us. The dataset comprises 78 fact-checked videos of electoral fraud claims made by voters during the 2022 Brazilian presidential election. We refer to this dataset as Election Integrity 22, shortened to EI22. In total, EI22 has 77 videos and 1997 segments, of which 78 are misinformative claims. The 77 videos are of varying lengths, come from voters' own recordings, and are unrelated to the videos on the BOL4Y dataset.

\section{Methodology}

We organize our methodology based on each dataset. Section \ref{meth_bol} discusses building the BOL4Y dataset and its related experiments, and Section \ref{meth_ei22} discusses building the EI22 dataset and the cross-dataset experiment. Section \ref{meth_classification} then details the classifiers we employed, and Section \ref{subsec:experimentalsetup} describes the setup that supports our proposed experiments.

\begin{figure*}
    \centering
    \includegraphics[width=0.8\textwidth]{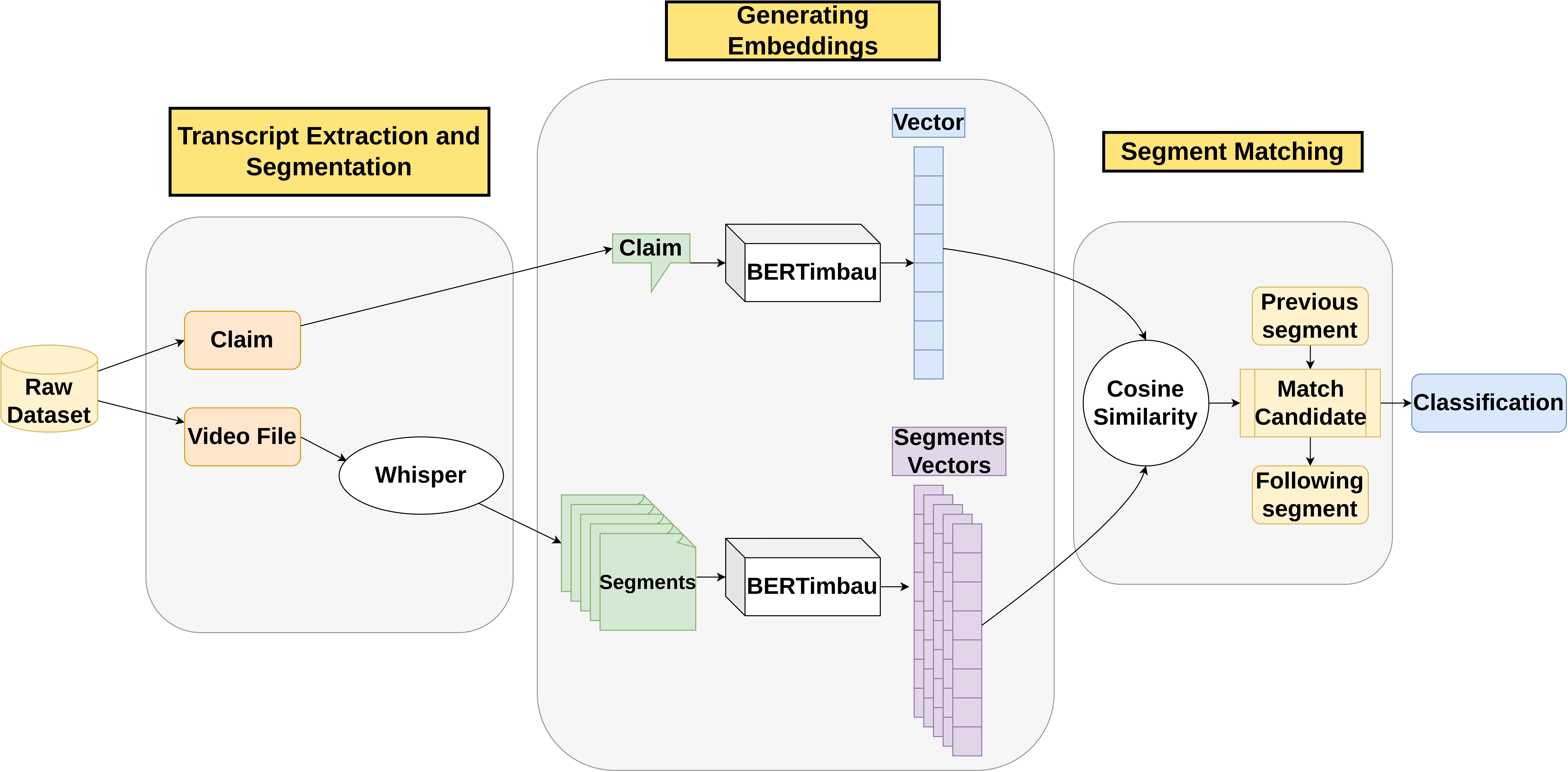}
    \caption{Overview of our methodology regarding the BOL4Y dataset}
    \label{fig:meth-overview}
\end{figure*}

\subsection{Building BOL4Y}\label{meth_bol}
Our methodology for building BOL4Y consists of the following steps:
1) \textbf{Transcript extraction and segmentation:} We normalize our dataset so that we convert each video to a textual transcript, which we segment into pieces; 
2) \textbf{Segment embeddings generation:} We convert the segmented textual data from the transcript into dense embeddings using a BERT-based model;
3) \textbf{Perform segment matching:} We semantically match the segments with the annotated false claims from AosFatos using the segment embeddings;
and 4) \textbf{Classification:} We perform a segment-level classification to identify segments sharing false information, essentially solving the misinformation spans detection task.
We present an overview of our methodology in Figure~\ref{fig:meth-overview}. 
Below, we elaborate on these steps and our experimental setup.

\subsubsection{Transcript Extraction \& Segmentation}


Our approach to misinformation span detection in videos leverages the transcriptions of videos' audios. To extract transcriptions from videos, we use OpenAI's Whisper model~\cite{whisper}, a state-of-the-art speech recognition model, on the audio of each video in our dataset. Whisper takes as input an audio file and generates a textual transcription. Although Whisper cannot provide word-level timestamps~\cite{githubdiscussion}, it can segment audio into transcribed segments (i.e., parts of the transcription) of at most 30-second windows.\footnote{As a limitation of Whisper, this window was not configurable at the time the dataset was transcribed.} We applied the OpenAI Whisper model to the 525 video files in our dataset and extracted their textual transcripts. Note that the transcripts provided by Escriba are already segmented by AosFatos' Escriba service.

\subsubsection{Generating Segment Embeddings}\label{sec:method-generating-embeddings}
Having converted our dataset into textual information (i.e., textual transcripts) divided into segments, our next step is to align the transcribed segments with manually annotated fact-checks provided by AosFatos. To this end, we use a state-of-the-art transformer-based model specifically trained and tailored for Brazilian Portuguese. 
Specifically, we use BERTimbau~\cite{bertimbau}, a BERT-based  model~\cite{devlin2018bert} that is pre-trained on the Brazilian Web as Corpus (BrWac)~\cite{bwac}, a large Portuguese corpus. The model was downloaded from the HuggingFace repository \footnote{https://huggingface.co/neuralmind/bert-base-portuguese-cased}, and we use the base model that yields embeddings of 768 dimensions. Moreover, we use the SentenceTransformers \cite{reimers-2019-sentence-bert} implementation to retrieve the embeddings from the mean pooling of the language model.
In a nutshell, BERTimbau takes as input the textual information included in a transcript segment and generates a dense vector representation (\emph{embedding}); these embeddings are the foundation for matching segments that share misinformation as they allow us to assess the similarity of transcript segments and fact-checked claims. 

\subsubsection{Performing Segment Matching}\label{sec:segmentmatching}

Here, we aim to identify the transcript segments that contain misinformation claims by comparing the segments with the fact-checked claims.
This is an integral part of our methodology as it allows us to create an annotated dataset of segments that share misinformation and segments that do not.
To achieve this, we perform the following procedure: After we use the BERTimbau model to extract an embedding for each false claim in our dataset (see \emph{Claim} field in Section~\ref{sec:dataset}), we compare each false claim embedding with all the segment embeddings (see Section~\ref{sec:method-generating-embeddings}) using cosine similarity. For each false claim, we consider the segment with the highest cosine similarity as the top candidate to be examined. This part allows finding the segments that potentially share misinformation, as they share textual similarities with the known false claims. Given that a false claim may span into multiple transcription segments, we also extract the segments before and after the segment with the highest cosine similarity for further examination.

After identifying segments that potentially share false claims, we perform a manual annotation process to verify that they are indeed sharing false claims.
We use the open-source text annotation tool Doccano\footnote{\url{https://github.com/doccano/doccano}} to speed up the annotation procedure.
For the annotation procedure, we focus on pairs of top candidate segments and false claims with a cosine similarity of 0.7 or higher.
We selected this threshold after manual examinations that showed that pairs with a cosine similarity of 0.7 or below were not semantically similar.
After applying this threshold and selecting all pairs of top candidate segments/false claims with cosine similarity higher than 0.7, we end up with 2,996 pairs that we annotate.
For each claim, we prompt the annotator to flag which of the three selected segments comprise the claim.
We choose a comprehensive approach and flag every segment that has at least one word that is part of the claim. 
We also add flags to i) signal if more segments are needed to capture the whole claim, ii) signal if there is a mistranscription (i.e., some words of one or more segments seem to be mistranscribed); iii) None of the segments shown match the fact-checked claim. 

We perform additional rounds of segment matching with the instances flagged as missing a part of the claim, adding more segments before and after the already flagged segments. Two annotators, authors of this work, matched 2,373 claims from the initial 6,685 claims listed by AosFatos, with the two annotators disagreeing only on 18 cases, which were discussed and removed from the dataset, resulting in 2,355 total segments with a 99.24\% agreement rate between annotators. Afterwards, for each matched claim, we concatenate the segments composing that claim into one and consider that concatenation as a positive example in further steps. The reason for merging these segments is to ensure that the full context of the claim is considered. In cases where a claim is spread across multiple segments, each segment on its own might not contain enough information to determine if it's misinformative. Note that these claims come from a subset of the initial set of downloaded videos: 430 videos out of the initial 525 and 108 out of the initial 121 transcriptions from Escriba, totaling 538 unique sources.



Finally, since our goal is to model our problem as a segment classification task, we need segments that do not share false claims (i.e., negative samples).
To do this, we treat all segments that are not matched or annotated as negative samples (i.e., segments that do not share false claims).
Using this approach, we end up with 336,855 segments that we treat as negative samples.

\subsection{Building EI22}\label{meth_ei22}

AosFatos provided us with EI22, which contained the videos and timestamps of the misinformative claims. We again employed Whisper, which transcribed the audio into segments. We then selected the segments that comprised the timestamps of the claims. EI22 consists of 77 videos and 78 claims in total.

\section{Resources Released and FAIR Principles}

The datasets are provided in textual form as .csv files, including 2,355 fact-checked claims for the BOL4Y dataset and 78 for the EI22 dataset. For each fact-check instance in each dataset, we release the corresponding audio from which the transcription was derived, the transcript itself, as well as the original videos (when available). As discussed previously, some sources for the BOL4Y were transcripts provided directly by AosFatos, with no video source available. Table \ref{tab:resources} summarizes the information released in each dataset. As further discussed in our methodology, each video transcript is split into segments. The total number of segments is shown in Table \ref{tab:resources}.

\begin{table}[ht!]
\centering
\resizebox{\columnwidth}{!}{%
\begin{tabular}{lccccc}
\hline
\textbf{Dataset}
& \makecell{\textbf{\# of}\\\textbf{Fact Checks}}
& \makecell{\textbf{Total}\\\textbf{segments}}
& \makecell{\textbf{Transcripts}\\\textbf{available}}
& \makecell{\textbf{Original videos}\\\textbf{available}} \\ \hline
BOL4Y & 2355 & 339K & 538 & 430\textsuperscript{\textdagger} \\
EI22  & 78   & 1997 & 77  & 77 \\ \hline
\end{tabular}}
\caption{Summary of data released}
\label{tab:resources}
\end{table}
\footnotetext[8]{\textsuperscript{\textdagger} As discussed, 118 transcripts come from Escriba, with no available video source}

We release the code used to build the dataset, as well as train models used in our experiments. For the BOL4Y dataset specifically, we also release the original HTML pages scraped as well as a parsed version in a .csv file. Additional details regarding fields of each .csv file are described in the appendix.  All datasets are available in a Zenodo repository (\url{https://zenodo.org/records/19097541}); due to Zenodo’s storage limitations, video and audio files are hosted separately on a HuggingFace repository, where files are available to the research community upon request (\url{https://huggingface.co/datasets/brenomatos/msd}). Code is released as a github repository.\footnote{https://github.com/brenomatos/msd} By hosting our dataset and code via Zenodo, Github and HuggingFace, we abide by FAIR principles by making assets findable, accessible, interoperable, and reusable.


\section{Baseline Classification}\label{meth_classification}

To investigate the feasibility of automatically detecting false segments in video transcripts, for our classification task, we employ two models pre-trained with Brazilian Portuguese: BERTimbau and PTT5. We use BERTimbau, which we already use for extracting segment embeddings, and PTT5~\cite{ptt5}, which was also pre-trained on the BrWac collection and is based on the T5 architecture~\cite{originalt5}. For each of these models, we use a classification head with a softmax activation that provides us, for each segment, a probability of the segment sharing false claims or not. Note that for the classification, we elected to use PTT5 in addition to BERTimbau to compare how the selection of the underlying Transformer architecture (i.e., encoder-only vs. encoder-decoder) affects the classification performance.

\subsection{Experimental Setup}\label{subsec:experimentalsetup}
Here, we provide more details on our experimental setup, including information about the dataset preparation, training and evaluation, and our sliding window experiments. 

\noindent \textbf{Dataset Preparation.} Our BOL4Y dataset is highly imbalanced: 2,355 positive instances (i.e., segments sharing false claims) and 336,885 negative instances (i.e., segments that do not share false claims).
This substantial class imbalance impacts the classification performance, hence, we evaluate the performance of the classification task using various configurations by randomly undersampling the negative examples in the training dataset.
In particular, we use the following ratios: 1-to-1 (i.e., balanced training set across classes), 1-to-10, 1-to-25, 1-to-50, 1-to-75, 1-to-100, and the full dataset (2.3K positive and 336K negative samples).


\noindent \textbf{Dataset Variations.} AosFatos published the list of claims on their website, but they may present editing by their journalists to correct grammatical errors or, in some cases, to add some context within brackets, which causes this version of the claim edited by the journalist to not necessarily be the same as what we find in the transcripts. Therefore, we have created an alternative version of the dataset in which we have replaced the false claims found in the transcripts with the version released by the journalist. Hence, we will refer to the variation of the dataset with the claims written by the journalist as the "Edited" version and the version with claims extracted from transcriptions as the "Original" version. For reference, we provide an example of what a claim looks like in the original and edited datasets in Table \ref{tab:editedvsoriginalexample}: note that the version from the edited dataset has context added in brackets. Given the polished nature of the edited dataset, we aim to provide insights into the challenges of working with transcriptions for misinformation span detection.
We aim to assess how the quality of the transcripts affects the classification performance when considering the misinformation span detection task.


\begin{table}[t]
\resizebox{\columnwidth}{!}{

\begin{tabular}{cc}
\toprule
\textbf{\begin{tabular}[c]{@{}c@{}}Dataset \\ Variation\end{tabular}} & \textbf{\begin{tabular}[c]{@{}c@{}}Bolsonaro's Claim\end{tabular}}           
\\\midrule 
\textbf{Original}                                                              & \begin{tabular}[c]{@{}c@{}}"He built three hydroelectric power plants abroad"\end{tabular}                                          \\ 
\textbf{Edited}                                                                & \begin{tabular}[c]{@{}c@{}}"He {[}Lula, Brazil's former president{]} \\ built three hydroelectric power plants abroad"\end{tabular} \\ \bottomrule
\end{tabular}
}
\caption{Example of claim in the original and edited datasets}
\label{tab:editedvsoriginalexample}
\end{table}


\noindent \textbf{Training and Evaluation.}
We use the HuggingFace implementations of the BERTimbau\footnote{https://huggingface.co/neuralmind/bert-base-portuguese-cased} and PTT5\footnote{https://huggingface.co/unicamp-dl/ptt5-base-portuguese-vocab} models, which we fine-tune for our dataset variations using Nvidia T4 GPUs. 
The HuggingFace implementations contain a classification head that produces the output prediction from the generated embeddings of the model.
We perform classification with 5-fold cross-validation. For each fold, we divide the dataset into five equal portions; three are used for training, one for validation, and one for testing. 
The validation set is used in an early-stopping approach, as we use the model from the epoch that best performed in the validation set.
We train the models for three epochs and use default parameters from their implementation.
Then, we assess the performance of classifiers and the impact of training set sizes on evaluation results. 
The undersampled variations also give us an insight into how classifiers can be implemented and used in the wild, as a bigger dataset also implicates using more resources to train models.
We follow the above procedure considering different undersampling ratios over the original and edited datasets.


 \noindent \textbf{Sliding Window Experiments.} We conduct experiments in a temporal manner to evaluate the real-world feasibility of detecting misinformation in future data. Specifically, we investigate whether models trained on past months' data can accurately predict subsequent months' misinformation. We perform two types of experiments: 1) fixed training and 2) expanding training windows. 

In the first setting, the training and test sets span fixed periods (6 months for training and one month for testing). We progressively move the testing window forward by one month. In the second experiment, the test window remains fixed for one month, but the training window expands with each iteration. During training, we use the most recent month as validation data for both settings. We train each variation for three epochs and select the model with the best performance on the validation set.
The temporal experiments aim to understand: If we train models on data from a given period in months, can we accurately predict misinformation in future months?


\noindent \textbf{Cross-dataset performance.} We also perform a cross-dataset test, training models with BOL4Y and testing on EI22. We train for three epochs with BERTimbau and PTT5 using multiple undersampling ratios.

Apart from releasing additional data (i.e., the EI22 dataset) for the task, we aim to provide insights into data representativeness of data and misinformative claims; we also strive to assess how models trained in one dataset perform when tested in another dataset of claims made by different speakers, further discussed in Section \ref{sec:classification-performance}, effectively showcasing the feasibility of the task in a real-world scenario.

\noindent \textbf{Limitations.} Next, we discuss some limitations of our methodology. First, although we use Whisper, a high-quality transcription model, audio transcriptions can still be noisy data and transcription models depend heavily on the audio quality to yield good results. Additionally, Whisper does the segmentation process automatically and on a sentence level. Currently, Word-level segmentation is not  supported in Whisper~\cite{githubdiscussion}. Some transcriptions come from Escriba, AosFatos' proprietary transcription service that does not disclose details on implementation. 

Finally, our data is focused only in the Brazilian context, which is restricted to Portuguese language. Representativeness is an important but challenging issue in any empirical study, as ours. We argue that the Brazilian context is relevant for the study of misinformation and our data covers a wide range of themes highly exploited by misinformation campaigns along four years~\cite{nyt2018benevenuto,datasetfake_icwsm}. For instance, our data includes Bolsonaro's livestreams organized periodically used to construct narratives along different topics that would favor the former Brazilian president. 





%% file: sections/results.tex
\section{Results}

This section presents the classification results and a temporal analysis of the classification performance.

\begin{table*}[h!]
\centering
{\fontsize{8}{10}\selectfont
\begin{tabular}{lccccccc}
\hline
                             & \textbf{\begin{tabular}[c]{@{}c@{}}BERTimbau\\ (Full)\end{tabular}} & \textbf{\begin{tabular}[c]{@{}c@{}}BERTimbau\\ (1-to-1)\end{tabular}} & \textbf{\begin{tabular}[c]{@{}c@{}}BERTimbau\\ (1-to-10)\end{tabular}} & \textbf{\begin{tabular}[c]{@{}c@{}}BERTimbau\\ (1-to-25)\end{tabular}} & \textbf{\begin{tabular}[c]{@{}c@{}}BERTimbau\\ (1-to-50)\end{tabular}} & \textbf{\begin{tabular}[c]{@{}c@{}}BERTimbau\\ (1-to-75)\end{tabular}} & \textbf{\begin{tabular}[c]{@{}c@{}}BERTimbau\\ (1-to-100)\end{tabular}} \\ \hline
\textbf{Balanced Accuracy}   & 0.55                                                                & 0.82                                                                  & 0.78                                                                   & 0.75                                                                   & 0.68                                                                   & 0.69                                                                   & 0.62                                                                    \\
\textbf{Macro F1}            & 0.56                                                                & 0.49                                                                  & 0.63                                                                   & 0.67                                                                   & 0.66                                                                   & 0.68                                                                   & 0.63                                                                    \\
\textbf{Precision (Class 1)} & 0.21                                                                & 0.09                                                                  & 0.24                                                                   & 0.35                                                                   & 0.38                                                                   & 0.43                                                                   & 0.35                                                                    \\
\textbf{Recall (Class 1)}    & 1.00                                                                & 0.75                                                                  & 0.94                                                                   & 0.97                                                                   & 0.99                                                                   & 0.99                                                                   & 1.00                                                                    \\ \hline
                             & \textbf{\begin{tabular}[c]{@{}c@{}}PTT5\\ (Full)\end{tabular}}      & \textbf{\begin{tabular}[c]{@{}c@{}}PTT5\\ (1-to-1)\end{tabular}}      & \textbf{\begin{tabular}[c]{@{}c@{}}PTT5\\ (1-to-10)\end{tabular}}      & \textbf{\begin{tabular}[c]{@{}c@{}}PTT5\\ (1-to-25)\end{tabular}}      & \textbf{\begin{tabular}[c]{@{}c@{}}PTT5\\ (1-to-50)\end{tabular}}      & \textbf{\begin{tabular}[c]{@{}c@{}}PTT5\\ (1-to-75)\end{tabular}}      & \textbf{\begin{tabular}[c]{@{}c@{}}PTT5\\ (1-to-100)\end{tabular}}      \\ \hline
\textbf{Balanced Accuracy}   & 0.54                                                                & 0.81                                                                  & 0.76                                                                   & 0.70                                                                   & 0.64                                                                   & 0.60                                                                   & 0.58                                                                    \\
\textbf{Macro F1}            & 0.54                                                                & 0.49                                                                  & 0.61                                                                   & 0.64                                                                   & 0.62                                                                   & 0.60                                                                   & 0.58                                                                    \\
\textbf{Precision (Class 1)} & 0.15                                                                & 0.08                                                                  & 0.20                                                                   & 0.30                                                                   & 0.29                                                                   & 0.28                                                                   & 0.27                                                                    \\
\textbf{Recall (Class 1)}    & 1.00                                                                & 0.76                                                                  & 0.94                                                                   & 0.97                                                                   & 0.99                                                                   & 0.99                                                                   & 1.00                                                                    \\ \hline
\end{tabular}}
\caption{Classification results for our dataset}
\label{tab:results_5fold}

\end{table*}

\begin{table*}[h!]
\centering
{\fontsize{8}{10}\selectfont
\begin{tabular}{lcccccc}
\hline
                             & \textbf{\begin{tabular}[c]{@{}c@{}}BERTimbau\\ (1-to-1)\end{tabular}} & \textbf{\begin{tabular}[c]{@{}c@{}}BERTimbau\\ (1-to-10)\end{tabular}} & \textbf{\begin{tabular}[c]{@{}c@{}}BERTimbau\\ (1-to-25)\end{tabular}} & \textbf{\begin{tabular}[c]{@{}c@{}}BERTimbau\\ (1-to-50)\end{tabular}} & \textbf{\begin{tabular}[c]{@{}c@{}}BERTimbau\\ (1-to-75)\end{tabular}} & \textbf{\begin{tabular}[c]{@{}c@{}}BERTimbau\\ (1-to-100)\end{tabular}} \\ \hline
\textbf{Balanced Accuracy}   & 0.91                                                                  & 0.92                                                                   & 0.88                                                                   & 0.85                                                                   & 0.85                                                                   & 0.81                                                                    \\
\textbf{Macro F1}            & 0.60                                                                  & 0.73                                                                   & 0.78                                                                   & 0.81                                                                   & 0.81                                                                   & 0.81                                                                    \\
\textbf{Precision (Class 1)} & 0.21                                                                  & 0.39                                                                   & 0.52                                                                   & 0.62                                                                   & 0.65                                                                   & 0.68                                                                    \\
\textbf{Recall (Class 1)}    & 0.87                                                                  & 0.97                                                                   & 0.98                                                                   & 0.99                                                                   & 0.99                                                                   & 1.00                                                                    \\ \hline
                             & \textbf{\begin{tabular}[c]{@{}c@{}}PTT5\\ (1-to-1)\end{tabular}}      & \textbf{\begin{tabular}[c]{@{}c@{}}PTT5\\ (1-to-10)\end{tabular}}      & \textbf{\begin{tabular}[c]{@{}c@{}}PTT5\\ (1-to-25)\end{tabular}}      & \textbf{\begin{tabular}[c]{@{}c@{}}PTT5\\ (1-to-50)\end{tabular}}      & \textbf{\begin{tabular}[c]{@{}c@{}}PTT5\\ (1-to-75)\end{tabular}}      & \textbf{\begin{tabular}[c]{@{}c@{}}PTT5\\ (1-to-100)\end{tabular}}      \\ \hline
\textbf{Balanced Accuracy}   & 0.90                                                                  & 0.90                                                                   & 0.88                                                                   & 0.81                                                                   & 0.80                                                                   & 0.77                                                                    \\
\textbf{Macro F1}            & 0.58                                                                  & 0.71                                                                   & 0.75                                                                   & 0.76                                                                   & 0.79                                                                   & 0.76                                                                    \\
\textbf{Precision (Class 1)} & 0.19                                                                  & 0.37                                                                   & 0.46                                                                   & 0.54                                                                   & 0.63                                                                   & 0.60                                                                    \\
\textbf{Recall (Class 1)}    & 0.85                                                                  & 0.97                                                                   & 0.98                                                                   & 0.99                                                                   & 1.00                                                                   & 0.99                                                                    \\ \hline
\end{tabular}}
\caption{Classification results for the Edited version of our dataset.}
\label{tab:results_journalist}

\end{table*}

\subsection{Classification Performance}\label{sec:classification-performance}

\subsubsection{Original Dataset}
Table \ref{tab:results_5fold} shows the results for our classifiers regarding all considered variations of our training dataset. Recall that, due to the considerable amount of data, we leverage undersampling variations of our dataset as our training set while maintaining the same test sets for all experiments. We consider six positive-to-negative example ratios (1-to-1, 1-to-10, 1-to-25, 1-to-50, 1-to-75, and 1-to-100) along the full dataset when undersampling our training set. We implement a 5-fold cross-validation approach and report average values on a video level, i.e., we compute metrics for every video in every fold and report average value for five folds. We compare results with the Macro F1 score due to class imbalance, along with class-balanced accuracy, and precision and recall for Class 1 (misinformation).

The BERTimbau classifier trained on the full version of our dataset is outperformed by all undersampled versions. The same happens for the PTT5 classifier trained on the full dataset. These results motivate us to exclude the full version of the dataset from further experiments due to its poor performance and high training time.
The BERTimbau-based classifiers match or outperform the PTT5 ones when comparing the same training sets regarding Macro F1. The BERTimbau-based classifier trained on the smallest training set (1-to-1 ratio) yields the best-balanced accuracy value, achieving a 0.82 score, although with poorer recall, precision, and Macro F1. Regarding Macro F1, BERT (1-to-75) yields the best performance overall, with a Macro F1 score of 0.68. 
We see a positive impact on performance when varying the undersampling ratio, with better results than training models with the full dataset. This shows that training models in a full dataset setting can be counterproductive in addition to being more costly. Overall, these results highlight that misinformation span detection is challenging, with modern classifiers based on state-of-the-art language models achieving an F1 score of up to 0.68.



\subsubsection{Edited Dataset}
We also propose an analysis of classification using an alternative version of our dataset where we consider the claims as edited by the journalist. To provide context, we initially performed a sentence-matching task to locate fact-checked claims within video transcriptions. To better understand the challenges of using transcriptions as input for classification, we have created an alternative version of the dataset. In this version, we replaced the transcribed claims (which served as positive examples) with the original claims as presented by AosFatos' journalists. These original claims are more refined and polished in comparison.

We see an increase in performance when using the edited version of the dataset (See Table \ref{tab:results_journalist}) when comparing models trained in datasets with different undersampling ratios, which showcases the difficulty of working with transcriptions, which can be noisy. Particularly for PTT5, the best-performing version is now the 1-to-75 undersampled version instead of 1-to-25, as shown in Table~\ref{tab:results_5fold}.






\subsection{Temporal Analyses}


\begin{figure*}[ht!]
\centering
\includegraphics[width=0.95\textwidth]{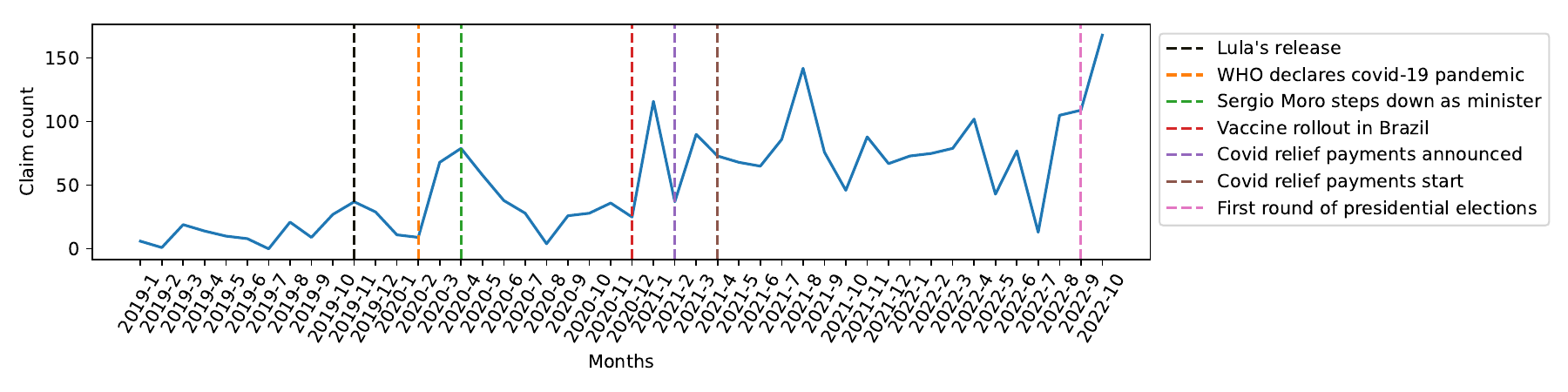}
\caption{Monthly sum of misinformation claims. Vertical lines signal important events during Bolsonaro's administration.}
\label{fig:monthly-sum}
\end{figure*}
 


\begin{figure*}[!ht]
\centering
\subfloat[BERTimbau 75 - 6 Month Training Period]{\includegraphics[width=0.9\textwidth]{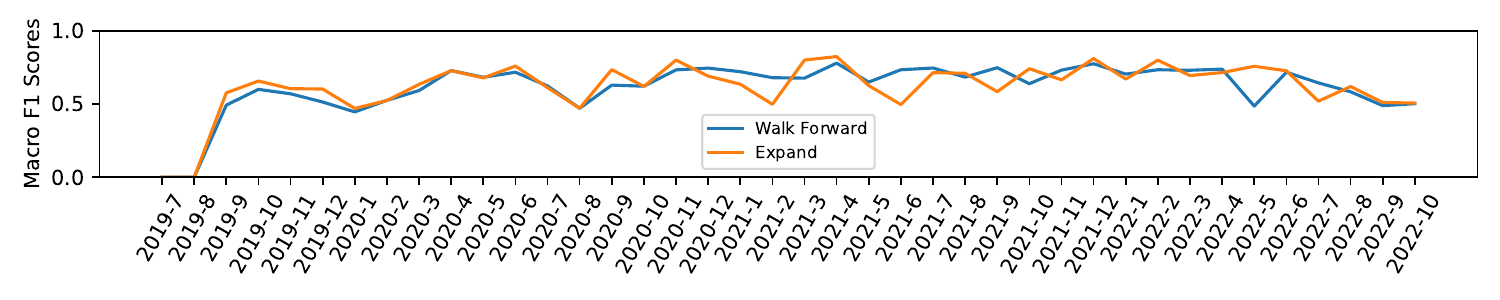}\label{fig:bert100temporal}}\hskip1ex
\subfloat[PTT5 25 - 6 Month Training Period]{\includegraphics[width=0.9\textwidth]{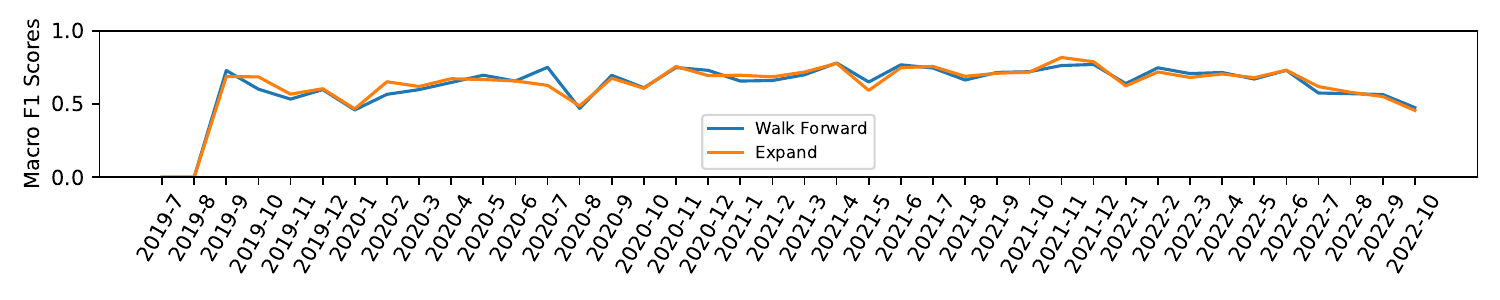}\label{fig:t510temporal}}
\caption{Temporal analysis of the performance of our classifiers.}\label{fig:temporal-results}
\end{figure*}

We conduct experiments where we partition the dataset by organizing the claims according to the specific months when Bolsonaro made them. We aim to gain insights into the practicality of deploying misinformation detection models in real-world scenarios where future data is inaccessible. This experiment will help us evaluate the robustness of our models in predicting and detecting future misinformation, focusing on the task of misinformation span detection.
We base our temporal analysis on the best-performing models in Table \ref{tab:results_5fold} regarding Macro F1 scores, namely BERT-75 and T5-25. Then, we propose two separate temporal studies for each: 1) a fixed training span of six months, hereafter referenced as \textbf{Walk-Forward} and 2) an increasing training span, hereafter referenced as \textbf{Expand}, starting with six months. 

Figure \ref{fig:monthly-sum} shows the distribution of Bolsonaro's false claims over time and important milestones of his presidency. Notably, false claims increased during the COVID-19 pandemic, starting to lower after the first quarter of 2022 and swiftly growing nearer to the presidential elections, when Bolsonaro faced his biggest political opponent, Brazil's then-former president, Lula.

We train all models for three epochs and, considering the last month of the training set as a validation set, choose the best version using early stopping. Note that due to the lack of claims in June 2019 (see Figure \ref{fig:monthly-sum}), we could not use it as a validation or test set, yielding null scores for June 2019 (test) and July 2019 (validation). In both settings ("Walk Forward" and "Expand"), we test models on the month chronologically after the month of the validation set. We consider the unedited dataset and report Macro F1 scores monthly in Figure \ref{fig:temporal-results}. Results show values ranging from 0.5 to 0.8, with the overall highest score on April 2021 (BERTimbau - Expand), the month after the start of Covid relief payments. For all settings, we observe a decrease in Macro F1 scores during the second semester of 2022, even for the "Expand" approaches, which are trained on all previous months. Overall, we also note that PTT5 yields more consistent performance across settings, generating similar results for "Walk Forward" and "Expand" in contrast to BERTimbau.

\subsubsection{Cross-dataset performance}

Table \ref{tab:cross-dataset} shows the cross-dataset experiment results. We trained models with BOL4Y and tested on EI22. We varied the undersampling ratio, achieving the best result (Macro F1 score of 0.72) with the 1-to-10 ratio for both models. Our results point to cross-dataset effectiveness, which is crucial in dealing with misinformation in a realistic setting.

\begin{table}[H]
\centering
\begin{tabular}{lccc}
\hline
                   & \textbf{1-to-1}  & \textbf{1-to-10} & \textbf{1-to-25}  \\ \hline
\textbf{BERTimbau} & 0.64             & 0.71             & 0.62              \\
\textbf{PTT5}      & 0.64             & 0.71             & 0.63              \\ \hline
                   & \textbf{1-to-50} & \textbf{1-to-75} & \textbf{1-to-100} \\ \hline
\textbf{BERTimbau} & 0.62             & 0.58             & 0.61              \\
\textbf{PTT5}      & 0.57             & 0.59             & 0.56              \\ \hline
\end{tabular}
\caption{Macro F1 scores for cross-dataset performance}
\label{tab:cross-dataset}
\end{table}

%% file: sections/conclusions.tex
\section{Concluding Discussion}


In this work, we present the first effort to explore the problem of misinformation span detection in videos. In addition to determining whether a video contains misinformation, we also identify the specific part (span) of the video where it occurs. We investigated multiple setups to assess the challenges related to effective misinformation span detection. We defined the first baseline for the task with an F1 score of 0.68, indicating the feasibility of the task. Furthermore, we built the first two datasets for misinformation span detection and made them available to the scientific community as one of the contributions of our work; our datasets provide completely novel data for a new, unexplored task. We also assessed cross-dataset performance, achieving an F1 score of 0.71 when training with the BOL4Y dataset and testing it on the EI22 dataset with both BERTimbau and PTT5; this points to effective detection despite misinformative claims coming from different sources. We also release all audios extracted from the video. We hope that the release of these datasets will open new avenues for research, enabling a deeper exploration of the problem through the lens of diverse machine learning solutions. Furthermore, we believe that our data construction methodology can be adopted by the community to generate similar resources, helping to address the scarcity of annotated data in this highly relevant domain. Future work will also explore the performance of LLMs and vision–language models for misinformation span detection, further expanding the set of available baselines. \looseness=-1


Additionally, we hope our work can motivate the development of novel applications to assist fact-checkers and reduce the time spent on misinformation detection in videos by pinpointing potential fact-checking points. Also, we argue that identifying the spans of misinformation within videos can assist social media operators in providing additional context to viewers when a false claim occurs. For instance, they can include warning labels with additional context regarding a false claim as an overlay on a video when a false claim is made. Finally, our work comes at a critical time for digital platforms. Initiatives like the Digital Services Act (DSA) regulation have emerged as significant steps forward in regulating digital spaces, aiming to ensure safer and more responsible online environments through effective content moderation. Such initiatives highlight the need for more robust automatic moderation tools, and we hope our work can improve these efforts. \looseness=-1

%% file: sections/future-work.tex

%% file: sections/ack.tex
\section*{Acknowledgments}


We thank our colleagues at AosFatos for sharing fact-checking data and providing initial feedback. This work was partially supported by INCT-IACiber (grant \#408432/2024-1), INCT-TILD-IAR (grant \#408490/2024-1), individual grants from CNPq, Fapemig and Kunumi Institute, and funding from Google Brazil.


%% file: sections/paper-checklist.tex
\subsection*{Paper Checklist}

\begin{enumerate}

\item For most authors...
\begin{enumerate}
    \item  Would answering this research question advance science without violating social contracts, such as violating privacy norms, perpetuating unfair profiling, exacerbating the socio-economic divide, or implying disrespect to societies or cultures?
    \answerYes{Yes}
  \item Do your main claims in the abstract and introduction accurately reflect the paper's contributions and scope?
    \answerYes{Yes}
   \item Do you clarify how the proposed methodological approach is appropriate for the claims made? 
    \answerYes{Yes}
   \item Do you clarify what are possible artifacts in the data used, given population-specific distributions?
    \answerYes{Yes}
  \item Did you describe the limitations of your work?
    \answerYes{Yes}
  \item Did you discuss any potential negative societal impacts of your work?
    \answerNo{No, we did not see any potentialnegative societal impacts}
      \item Did you discuss any potential misuse of your work?
    \answerNo{No, we did not see a potential for misuse}
    \item Did you describe steps taken to prevent or mitigate potential negative outcomes of the research, such as data and model documentation, data anonymization, responsible release, access control, and the reproducibility of findings?
    \answerYes{Yes}
  \item Have you read the ethics review guidelines and ensured that your paper conforms to them?
    \answerYes{Yes}
\end{enumerate}

\item Additionally, if your study involves hypotheses testing...
\begin{enumerate}
  \item Did you clearly state the assumptions underlying all theoretical results?
    \answerNA{NA}
  \item Have you provided justifications for all theoretical results?
    \answerNA{NA}
  \item Did you discuss competing hypotheses or theories that might challenge or complement your theoretical results?
    \answerNA{NA}
  \item Have you considered alternative mechanisms or explanations that might account for the same outcomes observed in your study?
    \answerNA{NA}
  \item Did you address potential biases or limitations in your theoretical framework?
    \answerNA{NA}
  \item Have you related your theoretical results to the existing literature in social science?
    \answerNA{NA}
  \item Did you discuss the implications of your theoretical results for policy, practice, or further research in the social science domain?
    \answerNA{NA}
\end{enumerate}

\item Additionally, if you are including theoretical proofs...
\begin{enumerate}
  \item Did you state the full set of assumptions of all theoretical results?
    \answerNA{NA}
	\item Did you include complete proofs of all theoretical results?
    \answerNA{NA}
\end{enumerate}

\item Additionally, if you ran machine learning experiments...
\begin{enumerate}
  \item Did you include the code, data, and instructions needed to reproduce the main experimental results (either in the supplemental material or as a URL)?
    \answerYes{Yes}
  \item Did you specify all the training details (e.g., data splits, hyperparameters, how they were chosen)?
    \answerYes{Yes}
     \item Did you report error bars (e.g., with respect to the random seed after running experiments multiple times)?
    \answerNA{NA}
	\item Did you include the total amount of compute and the type of resources used (e.g., type of GPUs, internal cluster, or cloud provider)?
    \answerYes{Yes}
     \item Do you justify how the proposed evaluation is sufficient and appropriate to the claims made? 
    \answerYes{Yes}
     \item Do you discuss what is ``the cost`` of misclassification and fault (in)tolerance?
    \answerNo{No}
  
\end{enumerate}

\item Additionally, if you are using existing assets (e.g., code, data, models) or curating/releasing new assets, \textbf{without compromising anonymity}...
\begin{enumerate}
  \item If your work uses existing assets, did you cite the creators?
    \answerYes{Yes}
  \item Did you mention the license of the assets?
    \answerNo{No, these
are defined on the original sources for the assets}
  \item Did you include any new assets in the supplemental material or as a URL?
    \answerYes{Yes}
  \item Did you discuss whether and how consent was obtained from people whose data you're using/curating?
    \answerNA{NA}
  \item Did you discuss whether the data you are using/curating contains personally identifiable information or offensive content?
    \answerNA{NA}
\item If you are curating or releasing new datasets, did you discuss how you intend to make your datasets FAIR (see \citet{fair})?
\answerYes{Yes, we released the datasets and associated data (transcripts, audio, videos, code, html pages.)}
\item If you are curating or releasing new datasets, did you create a Datasheet for the Dataset (see \citet{gebru2021datasheets})? 
\answerYes{Yes, in the Zenodo repository}
\end{enumerate}

\item Additionally, if you used crowdsourcing or conducted research with human subjects, \textbf{without compromising anonymity}...
\begin{enumerate}
  \item Did you include the full text of instructions given to participants and screenshots?
    \answerNA{NA}
  \item Did you describe any potential participant risks, with mentions of Institutional Review Board (IRB) approvals?
    \answerNA{NA}
  \item Did you include the estimated hourly wage paid to participants and the total amount spent on participant compensation?
    \answerNA{NA}
   \item Did you discuss how data is stored, shared, and deidentified?
   \answerNA{NA}
\end{enumerate}

\end{enumerate}

%% file: sections/appendix.tex
\appendix
\section*{Appendix A: Data Schema}

\Cref{tab:schema-dump,tab:schema-ei22,tab:schema-bol4y} show the description for each dataset

\begin{table*}[ht!]
\centering
\small
\begin{tabular}{@{}lp{0.55\linewidth}}
\toprule
\textbf{Field} & \textbf{Description} \\
\midrule
\texttt{title} & Title of the fact check \\
\texttt{date} & Posting date of the fact check \\
\texttt{aos\_fatos\_link} & URL to the original AosFatos page \\
\texttt{fact\_check} & Fact-checking paragraph \\
\texttt{topic\_pt} & Topic(s) of the false claim (Portuguese) \\
\texttt{source} & Origin of the claim (e.g., livestream, speech) \\
\texttt{source\_urls} & URL(s) to the content fact-checked \\
\texttt{repetition\_count} & Number of times the same claim was repeated \\
\texttt{year\_days\_pair} & Year/month/day(s) when the claim was made (incl.\ repetitions) \\
\texttt{page} & Crawl page where this fact check was found \\
\texttt{fact\_check\_id} & Fact-check identifier (join key with \texttt{BOL4Y}) \\
\bottomrule
\end{tabular}
\caption{Schema of \texttt{dump.csv}. Column names were translated to English for convenience.}
\label{tab:schema-dump}
\end{table*}

\begin{table*}[ht!]
\centering
\small
\begin{tabular}{@{}lp{0.55\linewidth}}
\toprule
\textbf{Field} & \textbf{Description} \\
\midrule
\texttt{fact\_check\_id} & Fact-check identifier \\
\texttt{file} & File source for this transcription \\
\texttt{transcription\_source} & Transcription system (\texttt{escriba} or \texttt{whisper}) \\
\texttt{transcription\_index} & Transcript segment id(s); may contain multiple ids due to concatenation \\
\texttt{transcription\_text} & Segment/claim text \\
\texttt{label} & \texttt{1} for misinformation, \texttt{0} for non-misinformation \\
\texttt{fixed\_date} & Date of the fact-check \\
\texttt{unique-check} & Identifier combining file + segment information \\
\bottomrule
\end{tabular}
\caption{Schema of \texttt{BOL4Y}. Each row corresponds to one transcript segment}
\label{tab:schema-bol4y}
\end{table*}

\begin{table*}[ht!]
\centering
\small
\begin{tabular}{@{}lp{0.55\linewidth}}
\toprule
\textbf{Field} & \textbf{Description} \\
\midrule
\texttt{file\_id} & File from which the segment was extracted \\
\texttt{transcript\_timestamp\_ids} &      Transcript segment id(s); may contain multiple ids due to concatenation \\
\texttt{transcript} & Full transcript for \texttt{file\_id} (only available for EI22 due to size constraints) \\
\texttt{minute\_timestamp} & Minute timestamp provided by AosFatos \\
\texttt{description} & Metadata provided by AosFatos \\
\texttt{label} & \texttt{1} for misinformation, \texttt{0} for non-misinformation \\
\texttt{segment\_text} & Segment/claim text \\
\bottomrule
\end{tabular}
\caption{Schema of \texttt{EI22}. Each row corresponds to one transcript segment.}
\label{tab:schema-ei22}
\end{table*}